# Training $L_1$-Regularized Models with Orthant-Wise Passive Descent Algorithms

**Jianqiao Wangni**
Tencent AI Lab
Shenzhen, China
zjnqha@gmail.com

## Abstract

The $L_1$-regularized models are widely used for sparse regression or classification tasks. In this paper, we propose the orthant-wise passive descent algorithm (OPDA) for optimizing $L_1$-regularized models, as an improved substitute of proximal algorithms, which are the standard tools for optimizing the models nowadays. OPDA uses a stochastic variance-reduced gradient (SVRG) to initialize the descent direction, then apply a novel alignment operator to encourage each element keeping the same sign after one iteration of update, so the parameter remains in the same orthant as before. It also explicitly suppresses the magnitude of each element to impose sparsity. The quasi-Newton update can be utilized to incorporate curvature information and accelerate the speed. We prove a linear convergence rate for OPDA on general smooth and strongly-convex loss functions. By conducting experiments on $L_1$-regularized logistic regression and convolutional neural networks, we show that OPDA outperforms state-of-the-art stochastic proximal algorithms, implying a wide range of applications in training sparse models.

## Introduction

The machine learning community has been favouring $L_1$-regularized models like logistic regression and linear regression to build robust applications with high dimensional data (Friedman, Hastie, and Tibshirani 2001)(Candes and Tao 2007)(Bach et al. 2012)(Bühlmann and Van De Geer 2011). To process the rapidly increasing scale of internet data, many algorithms were proposed to speed up the training process (Efron et al. 2004)(Zhang 2011)(Nesterov 2013). One of most representative optimization method is the proximal algorithm (Parikh, Boyd, and others 2014), which sequentially takes a gradient descent step and then solves a proximal problem on the current parameter.

When the data number is very large, the stochastic gradient descent algorithm (SGD) (Zhang 2004)(Bottou 2010)(Shamir and Zhang 2013), as opposed to batch algorithms, updates parameters by processing data mini-batches with a higher frequency and minimizes the loss function in expectation, being especially suitable for problems with large condition numbers. However, SGD generally needs a decreasing stepsize to reduce the variance of gradients, and only yields a sublinear convergence rate. Recently, some variance-reduced stochastic algorithms, such as SVRG (Johnson and Zhang 2013) and SAGA (Defazio, Bach, and Lacoste-Julien 2014), can converge without decreasing stepsizes and achieve linear convergence rates on smooth and strongly-convex problems; they can converge at similar rates on $L_1$-regularized problems if combined with proximal algorithms (Shalev-Shwartz and Zhang 2014)(Xiao and Zhang 2014)(Reddi et al. 2016).

Another promising line of research is about the (quasi) Newton algorithm, which has been popular for decades. It captures more curvature information than first-order algorithms do. It is theoretically guaranteed and practically proved to converge faster, especially when the data dimension is relatively low (Nocedal and Wright 1999). L-BFGS (Liu and Nocedal 1989) circumvents the inversion of Hessian matrices by matrix-vector multiplications, consuming less memory, fitting well with high dimensional data.

Due to the non-differentiability, the $L_1$-regularized sparse models can not directly benefit from the fast convergence of quasi-Newton algorithms. A representative adaptation of L-BFGS for the problem is the orthant-wise limited memory quasi-Newton method (OWL-QN)(Andrew and Gao 2007). OWL-QN fits this problem uniquely by generalizing L-BFGS and adopting three gradient alignment steps, which make the parameters remain in the same orthant after updates, and defeated other major algorithms on solving $L_1$-regularized logistic regression, in comparisons conducted by (Yu et al. 2010). Later improved OWL-QN algorithms (Gong and Ye 2015a)(Gong and Ye 2015b) take a hybrid approach of OWL-QN updates and proximal gradient descent steps, proving a theoretical convergence for a family of nonconvex models with nonconvex regularizations.

The success of SGD has led to some progresses toward stochastic quasi-Newton algorithms (SQN) that employ subsampled gradients as initializations (Schraudolph et al. 2007). The regularized SQN (Mokhtari and Ribeiro 2014) further added an identity matrix to the inverse Hessian matrix to guarantee the positive-definiteness. (Moritz, Nishihara, and Jordan 2016) proposed a linear convergent SQN with SVRG initializing descent directions.

Inspired by the aforementioned research, in this paper, we proposed an efficient but simple algorithm for $L_1$-regularized problems. We propose a generalization of the alignment operator from OWL-QN, and use SVRG for initializing the descent direction. We also present multiple strategies for calculating the quasi-Newton direction. The algorithm is evaluated on both convex and nonconvex problems, and the experiments demonstrate a significant improvement upon the proximal algorithms.

## Preliminaries

We study the regularized function $P(x)$ on $x \in \mathbb{R}^D$ as

$$P(x) = F(x) + R(x), \quad (1)$$

where $F$ is the average of $N$ loss functions, each of which depends on a data sample, and $R$ is the $L_1$ regularization,

$$F(x) = \frac{1}{N} \sum_{i=n}^{N} f_n(x), \quad R(x) = \lambda ||x||_1. \quad (2)$$

**Assumption 1.** Each loss function $f_n : \mathbb{R}^D \to \mathbb{R}$ is twice differentiable, $\mu$-strongly-convex, and has $L$-Lipschitz continuous gradient ($L$-smoothness), such that

$$\mu I_D \preceq \nabla^2 f_n(x) \preceq L I_D, \quad \forall 1 \leq n \leq N \quad (3)$$

where $||\cdot||$ is Euclidean norm, $0 < \mu \leq L$ and $I_D \in \mathbb{R}^{D \times D}$ is an identity matrix. Their finite average $F(x)$ also satisfies the smoothness and the strong convexity.

**Definition 2.** A proximal gradient algorithm for the problem in Eq.(1) sequentially finds a minimum on a quadratic expansion at $x_{k-1} \in \mathbb{R}^D$ in the $k$-th step,

$$x_k = \underset{x \in \mathbb{R}^D}{\operatorname{argmin}} \nabla F(x_{k-1})^\top x + \frac{1}{2\eta}||x - x_{k-1}||^2 + R(x), \quad (4)$$

where $\eta$ is the stepsize. This step can be written as

$$x_k = prox_{\eta R}(x_{k-1} - \eta \nabla F(x_{k-1})), \quad (5)$$

by a proximal operator

$$prox_{\eta R}(y) = \underset{x \in \mathbb{R}^D}{\operatorname{argmin}} \frac{1}{2}||x - y||^2 + \eta R(x). \quad (6)$$

For $L_1$ regularization $R(x) = \lambda ||x||_1$, there is

$$prox_{\eta R}(x) = \operatorname{sign}(x) \odot \max(|x| - \eta\lambda, 0), \quad (7)$$

where $\odot$ is the element-wise product.

**Orthant-Wise Quasi Newton Method (OWL-QN).** The algorithm (Andrew and Gao 2007) restricts the updated parameter to be within certain orthants to keep differentiability of the $L_1$-regularized problem. We denote the sign function $\sigma()$ as follows: $\sigma(x_i) = 1$, if $x_i > 0$; $\sigma(x_i) = -1$ if $x_i < 0$ and $\sigma(x_i) = 0$ for otherwise. The alignment operator $\pi : \mathbb{R}^D \to \mathbb{R}^D$ is defined by per element as

$$\pi_i(x; y) = \begin{cases} x_i, & \text{if } \sigma(x_i) = \sigma(y_i), \\ 0, & \text{otherwise}, \end{cases} \quad (8)$$

where $y \in \mathbb{R}^D$ provides a reference orthant and $y_i$ is the $i$-th element of $y$. The alignment operator enforces two elements to have the same sign, so the two vectors are in the same orthant. For notation simplicity, we define an element-wise operator $\psi : \mathbb{R}^D \to \mathbb{R}^D$ as

$$\psi_i(v; x; \lambda) = \begin{cases} v_i + \lambda, & \text{if } x_i > 0 \\ v_i - \lambda, & \text{if } x_i < 0 \\ v_i + \lambda, & \text{if } x_i = 0, v_i + \lambda < 0 \\ v_i - \lambda, & \text{if } x_i = 0, v_i - \lambda > 0 \\ 0, & \text{otherwise}. \end{cases} \quad (9)$$

OWL-QN aligns the pseudo-gradient $\Diamond F(x)$ based on the current gradient $\nabla F(x)$ and the parameter $x$,

$$\Diamond F(x) = \psi(\nabla F(x); x; \lambda). \quad (10)$$

Then, by minimizing an approximated quadratic expansion at point $x_{k-1}$ of Eq.(1), OWL-QN finds a direction $d_k$ as

$$d_k = -\underset{d \in \mathbb{R}^D}{\operatorname{argmin}} F(x_{k-1}) + \Diamond F(x_{k-1})^\top d + d^\top B_k d/2$$
$$= H_k \Diamond F(x_{k-1}),$$

where $B_k$ is an approximated Hessian matrix at $x = x_{k-1}$ and $H_k = B_k^{-1}$. OWL-QN applies $H_k$ to obtain a quasi-Newton direction $d_k$, then takes the second alignment to obtain a direction $p_k$ which is orthant-wise equal to $\Diamond F(x)$,

$$p_k = \pi(d_k; \Diamond F(x^k)). \quad (11)$$

After this step, OWL-QN makes the third alignment, which explicitly restricts the updated parameter $x_{k-1} - \alpha_k p_k$ to be in the same orthant with $x_{k-1}$, as

$$x_k = \pi(x_{k-1} - \alpha_k p_k; x_{k-1}), \quad (12)$$

where the optimal stepsize $\alpha_k$ is obtained by line-searching.

## The Proposed Algorithm

Although being practically efficient, the aforementioned OWL-QN can certainly be further modified for possible improvement, for examples, by eliminating the line-search procedure, or using subsampled gradients instead of accurate but costly full gradients. Another promising research that attracts us is the stochastic variance-reduced gradient algorithm (SVRG)(Johnson and Zhang 2013), which adds a full gradient on a reference point to the subsampled gradient, and balances the gradient expection by substracting a subsampled reference gradient evaluated on the same subset. It converges well with a constant stepsize, and achieves a linear convergent rate for strongly-convex case. Inspired by SVRG and OWL-QN, we develop an improved algorithm that specializes in optimizing $L_1$-regularized problems. It combines the variance reduction technique of SVRG with the alignment operator of OWL-QN, but with a relative passive orthant-wise restriction on the parameter. We refer to it as the orthant-wise passive descent algorithm (OPDA).

**Definition 3.** A subsampled loss function $f_k$ in the $k$-th step is evaluated on a subset $S_k$, that

$$f_k(x) = \frac{1}{|S_k|} \sum_{n \in S_k} f_n(x), \quad \text{where } S_k \subset \{1, ..., N\}. \quad (13)$$

The stochastic variance-reduced gradient (SVRG) $v_k$ is as

$$v_k = \nabla f_k(x_{k-1}) - \nabla f_k(\tilde{x}) + \nabla F(\tilde{x}), \quad (14)$$

where $\tilde{x} = x_{k'}$ is a reference point obtained in a previous iteration $k'(k' < k)$, where the full gradient $\nabla F(\tilde{x})$ is calculated.

A combination of OWL-QN and SVRG seems simple, this, however, raises many non-trivial difficulties for us to address. The alignment operation sabotages the convergence property, being the main reason for OWL-QN not easy to prove within a decade.

**Remark 1.** For a modified OWL-QN that uses $d_k = H_k v_k$ as the descent direction, where $v_k$ is a stochastic gradient, we define the actual descent direction in the $k$-th step as $q_k$, then

$$\mathbb{E}[P(x_k)] = \mathbb{E}[P(x_{k-1} - \eta q_k)] \quad (15)$$
$$\leq F(x_{k-1}) - \eta \nabla F(x_{k-1})^\top q_k + \frac{\eta^2}{2}\mathbb{E}[||q_k||^2] + R(x_k).$$

The key to establish an optimizer that converges fast both in theory and application is to make $q_k$ as an unbiased estimate of $\partial P(x_{k-1})$, and to control the second-order moment $\mathbb{E}[||q_k||^2]$. OWL-QN suggests that

$$p_k = \pi(H_k(v_k + \partial R(x_{k-1})), v_k + \partial R(x_{k-1})),$$
$$q_k = (x_{k-1} - \pi(x_{k-1} - \eta p_k, x_{k-1}))/\eta.$$

As the algorithm approaches a stationary point, $v_k$ approaches the true gradient, as $v_k \to \nabla F(x_{k-1})$, supposing that $0 < \gamma, \gamma I_D \preceq H_k$ is positive-definite, and denoting $\bar{v}_k = \partial P(x_{k-1})$, then the key proposition of OWL-QN (Andrew and Gao 2007)

$$\bar{v}_k^\top q_k \geq \bar{v}_k^\top p_k \geq \bar{v}_k^\top H_k \bar{v}_k \geq \gamma ||\bar{v}_k||^2, \quad (16)$$

which is important for bounding the second term in Eq.(15), does not hold for the case.

The reasons for such a difficulty were originally discussed in (Gong and Ye 2015b), and their solution was to add proximal gradient descent steps to ensure the convergence. The analysis encourages us to modify the gradient alignment operation to ensure a straightforward convergence. To develop our algorithm, we propose the following propositions to define the aligned gradients, descent directions and orthant-wise updates, which are useful to describe OPDA.

**Proposition 4.** OPDA uses SVRG $v_k$ from Eq.(14) as the initializing descent direction, and uses the following pseudo-gradient to provide a reference orthant for $v_k$,

$$\Diamond f_k(x_{k-1}) = \psi(\nabla f_k(x_{k-1}), x_{k-1}, \lambda), \quad (17)$$

where $\lambda$ is the regularization parameter in Eq.(1).

**Remark 2.** Although a more straightforward choice is to set $v_k = \Diamond f_k(x_{k-1})$, but this will introduce a non-decreasing variance and compromise the convergence rate.

Next, we take the second alignment step, using the pseudo-gradient $\Diamond f_k(x_{k-1})$ as a reference point. This introduces a much smaller bias to $v_k$, since the two vectors are random with respect to each other.

**Proposition 5.** The descent direction $p_k$ is defined as

$$p_k = \pi(H_k v_k, \Diamond f_k(x_{k-1})), \quad (18)$$

where $H_k$ can be obtained from quasi-Newton methods or simply setting to be an identity matrix (first-order).

The aforementioned calculation does not explicitly involve the partial derivative of $R(x)$, except for the alignment reference, since we are avoiding additional variances. To make the solution path to be sparse, we introduce a novel alignment operator to encourage zero elements.

**Proposition 6.** The orthant-wise alignment operator $\phi$ is defined as

$$\phi_i(x; y; \varrho) = \begin{cases} 0, & \text{if } \sigma(x_i)\sigma(y_i) = -1 \text{ or } |x_i| < \varrho \\ x_i - \varrho \operatorname{sign}(x_i), & \text{otherwise}. \end{cases}$$

**Remark 3.** It is a generalization of $\pi(\cdot)$ operator with a flexible parameter $\varrho$. it reduces the absolute values of large elements and forces small elements to be zero.

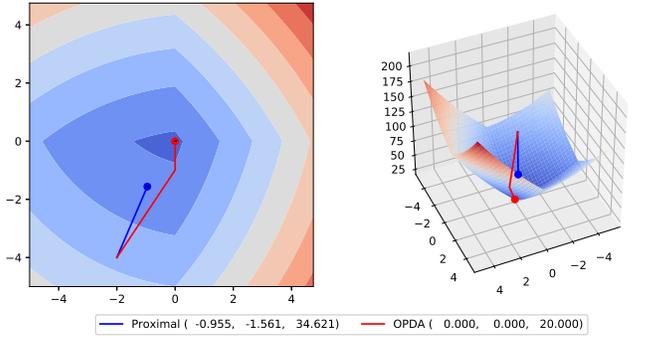

Figure 1: $P(x) = (x_1 + 4)^2 + (x_1 + 2)^2 + 10||x||_1$.

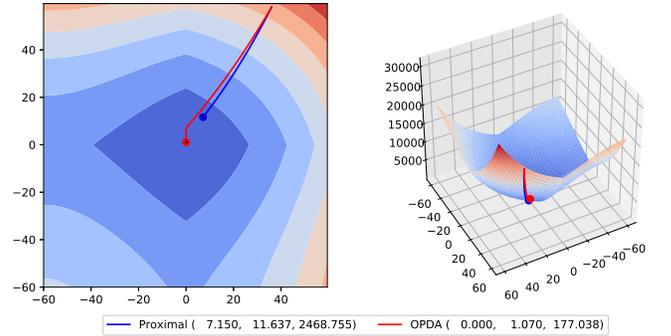

Figure 2: OPDA-FM and a proximal algorithm optimize $P(x) = (x_1 + 4)^2 + (x_1 + 2)^2 + 0.1x_1 x_2 + 0.02x_1^3 + 0.02(x_2 + 12)^3 + 100||x||_1$. (Two algorithms have the same stepsize and maximum iteration)

**Proposition 7.** OPDA updates the parameter and aligns it, based on the aligned direction $p_k$ scaled with a stepsize $\eta$,

$$x_k = \phi(x_{k-1} - \eta p_k; x_{k-1}; \eta\lambda),$$
$$q_k = (x_{k-1} - x_k)/\eta.$$

**Remark 4.** Different from OWL-QN, OPDA allows the update of any element of $x_{k-1}$ from zero to any orthant validated. Each element $p_i, i \in [D]$ of the descent direction $p$, is forced to zero if and only if $\sigma(p_i)\sigma(v_i) = -1$. OPDA drives the parameter passively across one orthant to another.

In conclusion, OPDA sequentially calculates the gradient $v$, the reference orthant $\sigma(\diamondsuit f(x))$, the direction $p$ and the actual update $q$, as the propositions above. We describe the whole framework in Algorithm 1. Throughout the following paper, we refer to OPDA that $d_k = v_k$ as OPDA-FM (the first-order method), refer to the one that $d_k = H_k v_k$ as OPDA-QN (quasi-Newton). To visualize a comparison, we plot the optimization trajectories of OPDA against a proximal algorithm on a synthetic function in Figures 1&2.

---

**Algorithm 1** Orthant-Wise Passive Descent Algorithms

**Input** $x_0 \in \mathbb{R}^D$, stepsize $\eta > 0$, and reference points update frequency $m$.
Initialize $k = 0$, $H_0 = I$, $t = 0$.
**repeat**
  Compute a full gradient $\nabla F(\tilde{x}_t)$ on the reference point.
  **repeat**
    Sample a random mini-batch $\mathcal{S}_k \subset \{1, \cdots, N\}$.
    Align the gradient $\diamondsuit f_k(x_{k-1})$ as Eq.(17).
    Compute the subsampled gradient $v_k$ as Eq.(14).
    **Choice 1 (OPDA-QN):** Compute the direction $d_k = H_k v_k$ via Algorithm 2 or via L-BFGS as Eq.(21).
    **Choice 2 (OPDA-FM):** Set the direction $d_k = -v_k$.
    Align the direction $p_k = \pi(d_k; \diamondsuit f_k(x))$.
    Align the update $x_k = \phi(x_{k-1} - \eta p_k; x_{k-1}; \eta \lambda)$. Set $k = k + 1$.
  **until** $k\%m == 0$
  Choice 1: Set $\tilde{x}_{t+1} = \frac{1}{m} \sum_{j=k-m}^{k} x_j$
  Choice 2: Set $\tilde{x}_{t+1} = x_j$, where $j$ is selected uniformly at random from $[m] = \{k-m, k-m+1, \cdots k\}$.
  Update Hessian matrix $H_k$ as Eq (21), and set $t = t+1$.
**until** Reaching maximum outer iterations $t = T$, or converging.

---

## Quasi-Newton Updates

For calculating the direction $d = Hv$, we can adopt the method of L-BFGS, where the inverse Hessian approximation $H_k$ is constructed from the curvature pairs $(s_j, y_j)$,

$$s_j = x_j - x_{j-1}, \quad y_j = \nabla F(x_j) - \nabla F(x_{j-1}) \quad (19)$$

for $j \leq k$. Denoting $\rho_j = 1/s_j^\top y_j$, initializing

$$H_k^{k-M} = (s_r^\top y_r / ||y_r||^2) I_D, \quad (20)$$

then OWL-QN recursively computes

$$H_k^j = (I - \rho_j s_j y_j^\top)^\top H_k^{j-1} (I - \rho_j s_j y_j^\top) + \rho_j s_j s_j^\top \quad (21)$$

for $k - M + 1 \leq j \leq k$, and assigns $H_k \leftarrow H_k^k$ as L-BFGS does. We can also incorporate a block BFGS method (Gower, Goldfarb, and Richtárik 2016) to accelerate calculating the direction $Hv$. This method is derived from L-BFGS, and it solves a linear system which is based on $H_{k-1}$ to obtain $H_k$,

$$H_k = \underset{H \in \mathbb{R}^{D \times D}}{\text{argmin}} ||(H - H_{k-1})\nabla^2 f_k(x_{k-1})||_F^2, \quad (22)$$

$$s.t. \quad H\nabla^2 f_k(x_{k-1})\Xi_k = \Xi_k; H = H^\top, \quad (23)$$

where $||\cdot||_F$ is the Frobenius norm. The matrix $\Xi_k$ hereby introduces randomness into the algorithm, acting as a sketching function. The system has a closed form solution as

$$\Delta_k = (\Xi_k^\top Y_k)^{-1}, \quad Y_k = \nabla^2 f_k(x_{k-1})\Xi_k,$$
$$H_k = \Xi_k \Delta_k \Xi_k^\top + (I - \Xi_k \Delta_k Y_k^\top) H_{k-1} (I - Y_k \Delta_k \Xi_k).$$

There are several kinds of random sketching strategies (Gower, Goldfarb, and Richtárik 2016), as follows: a) $\Xi_k$ is an identity matrix; b) each element of $\Xi_k$ is i.i.d sampled from a Gaussian distribution; c) stacks previous unaligned direction vectors as $\Xi_k = [d_{k+1-M}^\top, \cdots, d_k^\top]$.

We describe the matrix-vector multiplication for block BFGS in Algorithm 2. The update frequency of the reference point is indicated by $m$, which also indicates the times of evaluating subsampled gradients with the same full gradient $\nabla F(\tilde{x})$. The inverse Hessian $H_k$ is updated by every $M$ iterations, and uses $M$ saved $(s_j, y_j)$ curvature pairs (L-BFGS) (Byrd et al. 2016) or $(\Xi_j, Y_j, \Delta_j)$ curvature triples (BL-BFGS)(Gower, Goldfarb, and Richtárik 2016). Algorithm 2 consumes $M(4D + 2r)r$ arithmetic operations in total. Its cost is approximately $\mathcal{O}(D^{3/2})$ following the setting of (Gower, Goldfarb, and Richtárik 2016) that $r \leq \sqrt{D}$. The calculation of $\Xi_k^\top Y_k$ and the Cholesky factorization results in additional $\mathcal{O}(r^2 D)$ plus $\mathcal{O}(r^3)$ operations. The gradient alignment only consumes $\mathcal{O}(D)$ operations, which are negligible compared to the major costs.

---

**Algorithm 2** Block L-BFGS update

**Input** $v_k \in \mathbb{R}^D$, $\Xi_j, Y_j \in \mathbb{R}^{D \times r}$ and $\Delta_j \in \mathbb{R}^{r \times r}$ from Algorithm 1, for $j \in \{k+1-M, \cdots, k\}$.
Sample a matrix $\Xi_k \in \mathbb{R}^{D \times r}$ that $rank(\Xi_k) = r$.
Compute $Y_k = \nabla^2 f(x_k)\Xi_k$ and $\Xi_k^\top Y_k$.
Compute $\Delta_k = (D_k^\top Y_k)^{-1}$ by Cholesky factorization.
**Initialize** $v' = v_k, j = k$,
Repeat $\alpha_j = \Delta_j \Xi_j^\top v'$, $v' = v' - Y_j v'$, $j = j - 1$, until $j = k - M + 1$.
Repeat $\beta_j = \Delta_j Y_j^\top v'$, $v' = v' + \Xi_j(\alpha_j - \beta_j)$, $j = j + 1$, until $j = k$.
**Output:** $v' = H_k v_k$.

---

## Convergence Analysis

The difficulty of proving the convergence of an stochastic orthant-wise algorithm mainly rises from the two points: a), (for OPDA-QN) controlling the variance of the descent direction based on an inaccurate gradient. b) dealing with inconsistent orthant-wise constraints from alignment operators. The former issue is shared by all previous stochastic quasi-Newton (SQN) methods (Byrd et al. 2016)(Moritz, Nishihara, and Jordan 2016)(Gower, Goldfarb, and Richtárik 2016)(Luo et al. 2016)(Wang et al.

2017), even with smooth functions. Until now, the variance of the SQN descent direction can only be loosely bounded, therefore they there is no better convergence rate the best first-order methods. Although this drawback is critical, it is not in the scope of this paper. In this section, we follow analyses from (Xiao and Zhang 2014) (Reddi et al. 2016). First, we define the suboptimality function $Q$ as

$$Q(x) = P(x) - P(x_\star), \quad \text{where} \quad x_\star = \arg\min_x P(x).$$

Then we introduce several lemmas before proceeding to the main theorems.

**Lemma 8.** For OPDA with Assumption 1 holding, then
$$\mathbb{E}[||\nabla f_k(x) - \nabla f_k(x_\star)||^2] \leq 2LQ(x), \quad (24)$$

**Lemma 9.** For OPDA with Assumption 1 holding, we set $v_k$ as Eq.(14), then
$$\mathbb{E}[||v_k - \nabla F(x_{k-1})||^2] \leq 4L[Q(x_{k-1}) + Q(\tilde{x})]. \quad (25)$$

**Lemma 10.** For any convex function $R$ on $\mathbb{R}^D$, and $x, y \in \mathbb{R}^D$, it holds that
$$||prox_R(x) - prox_R(y)|| \leq ||x - y||. \quad (26)$$

**Lemma 11.** Suppose that Assumption 1 holds and each loss function is $\mu$-strongly-convex. Then there exists $\gamma I \preceq H_k \preceq \Gamma I, \forall k \geq 1$ for OPDA-QN where $0 < \gamma < \Gamma$.

**Remark 5.** For OPDA-FM, there ia a constant $\gamma = \Gamma = 1$ since $H_k = I$. The bound for L-BFGS update can be found in [(Moritz, Nishihara, and Jordan 2016), Lemma 4] that
$$1/((D+M)L) \leq \gamma, \quad \Gamma \leq ((D+M)L)^{D+M-1}/\mu \quad (27)$$
and the bound for block L-BFGS update can be found in (Gower, Goldfarb, and Richtárik 2016), Lemma 1] that
$$1/(1+ML) \leq \gamma, \Gamma \leq (1+\sqrt{\alpha})(1+1/(2\sqrt{\alpha}\mu+\alpha\mu)). \quad (28)$$
where $\alpha = (1+\sqrt{L/\mu})^2$. For nonconvex problems, we may ajust the quasi-Newton methods as (Wang et al. 2017). Since we use the unaligned gradient during the BFGS update, this lemma still holds in our case, giving us a method of bounding the descent direction variance.

**Lemma 12.** For convex function $R$ on $\mathbb{R}^D$, and $x, y \in \mathbb{R}^D$, $||\phi(x;z;\eta\lambda) - \phi(y;z;\eta\lambda)|| \leq ||x-y||$.

**Remark 6.** This is the non-expansiveness of OPDA based on a fixed reference orthant, resembling the property of proximal mapping in [(Rockafellar 2015) section 31].

**Lemma 13.** For any regularized function $P(x) = F(x) + R(x)$, where $F(x)$ is $\mu$-strongly-convex and it has $L$-Lipschitz continuous gradient, $R(x)$ is convex, we define
$$x^- = x - \eta p, x^+ = \phi(x^-; x; \eta\lambda), q = \frac{1}{\eta}(x - x^+) \quad (29)$$
and define $p \in \mathbb{R}^D, g \in \mathbb{R}^D, \Delta \in \mathbb{R}^D$ by
$$g_i = \begin{cases} q_i, (\sigma(x_i^-)\sigma(x_i) = -1) \\ p_i, (\sigma(x_i^-)\sigma(x_i) \neq -1), \end{cases} \Delta = g - \nabla F(x), \quad (30)$$
also define $\eta$ as the stepsize that $0 < \eta \leq 1/L$, then for any $y \in \mathbb{R}^D, x \in \mathbb{R}^D$ and $x^+ \in \mathbb{R}^D$, we have
$$P(y) \geq P(x^+) + q^\top(y-x) + \frac{\eta}{2}||q||^2$$
$$+ \frac{\mu}{2}||y-x||^2 + \Delta^\top(x^+ - y). \quad (31)$$

**Lemma 14.** Under the condition in Lemma 13, there is $||g|| \leq ||p||$ and $prox_{\eta R}(x - \eta g) = x^+$.

**Remark 7.** One can see that by viewing OPDA as a proximal algorithm, the second-order moment of the aligned direction $g$ is smaller than $p$.

Suppose that the dataset is normalized, and each element has an expectation of zero, then the optimization trajectory $\{x_k\}_{k=0}^\infty$ is uniformly distributed in all orthants, therefore we have the following reasonable assumption.

**Assumption 15.** Suppose $x \in \mathbb{R}^D$ is a random point from the optimization trajectory of OPDA, each element of $x$ has the same probability being positive or negative. For data samples $S_k$, the expectation of corresponding gradients is zero,
$$\mathbb{E}_{S_k}[\mathbb{E}_x[\nabla_i f_k(x)]] = \mathbb{E}_x[\nabla_i F(x)] = 0, \quad \forall i \in [D]. \quad (32)$$

**Theorem 16.** Suppose that Assumption 1 holds for each loss function $f_n$, then OPDA-FM in Algorithm 1 proceeds with a stepsize $0 < \eta < 1/(6L)$, the reference points $\{\tilde{x}_t\}_{t=1}^\infty$ converge to the global optima $x_\star$ in expectation, with a linear convergence rate, as
$$\mathbb{E}[Q(\tilde{x}_t)] \leq \left(\frac{2 + 8\mu L\eta^2(m+1)}{2\mu\eta(1-6L\eta)m}\right)^t Q(\tilde{x}_0). \quad (33)$$

**Corollary 17.** Suppose that Assumption 1 holds for each loss function $f_n$, then for OPDA-FM in Algorithm 1, by setting $\eta = \theta/L$, there is
$$\rho \approx \frac{L}{\mu\theta(1-6\theta)m} + \frac{4\theta}{(1-6\theta)}, \quad \mathbb{E}[Q(\tilde{x}_t)] \leq \rho^t \mathbb{E}[Q(\tilde{x}_0)],$$
if under appropriate settings that $m = \mathcal{O}(L/\mu)$ and $\theta$ is sufficiently small that $0 < \rho < 1$, then the iteration-complexity of attaining an $\epsilon$-accurate suboptimum is
$$\mathcal{O}\left((N + \frac{L}{\mu})\log(\frac{1}{\epsilon})\right).$$

**Remark 8.** The convergence rate of OPDA-FM for strongly-convex function resembles that of Proximal-SVRG (Xiao and Zhang 2014). This is due to that the main techniques used in the analyses are both non-expansiveness.

## Numerical Experiments
### Logistic Regression
First, we implement OPDA in MATLAB, based on the code generously provided by the authors of (Gower, Goldfarb, and Richtárik 2016). We verify the algorithm's efficiency by logistic regression with $L_2$ and $L_1$ regularizations for binary classification task. The objective function is
$$P(x) = \frac{1}{N}\sum_n -\log[1 + \exp(-a_n^\top x b_n)] + \lambda_2||x||_2^2$$
$$+ \lambda||x||_1, \quad a_n \in \mathbb{R}^D, \quad b_n \in \{-1, 1\}. \quad (34)$$

We use datasets from (Chang and Lin 2011), including **covtype** ($N = 581K, D = 54$) and **rcv1** ($N = 20K, D = 47K$). The regularization coefficients $\lambda_1$ and $\lambda_2$ are noted with the figures. If not noted, the default $L_1$ regularization is set to be $\lambda_1 = 10^{-5} \sqrt[4]{D}/\sqrt[4]{N}$ for rcv1 data, and $\lambda_1 = 10^{-5}$ for covtype; the default $L_2$ regularization for both datasets is set to be $\lambda_2 = 1/N$. We compare our algorithm (with a prefix **OPDA**) with the Proximal-SVRG algorithm (with a prefix **PrSVRG**). The stepsize $\eta$ is noted after the prefix.

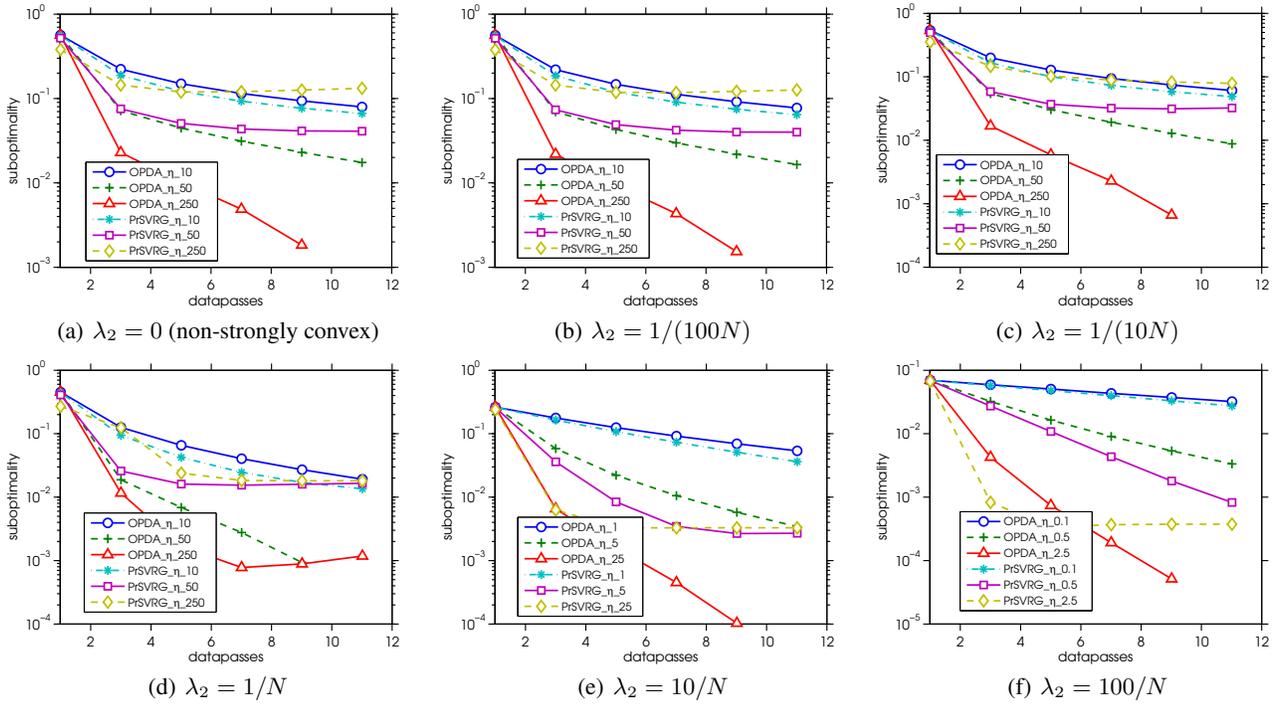

Figure 3: Performance of OPDA-FM for different $L_2$ regularization on rcv1 dataset.

For OPDA and Proximal-SVRG, we use the same sub-sampling size $|\mathcal{S}_k| = \sqrt{N}$ for the gradient and Hessian update (for OPDA-QN). The stepsize is grid-searched for an optimum, whose nearby stepsizes are also tested and plotted. We set each outer iteration to consist of $m = N/|\mathcal{S}_k|$ times of inner iteration, during which both OPDA and Proximal-SVRG fully scan over the dataset before recalculating the full gradient and updating the reference point. We plot the convergence of OPDA-FM for strongly-convex problems (with $L_2$ and $L_1$ regularization) in [Figure 3: (bcdef)], and a non-strongly-convex problem (without $L_2$ regularization) in [Figure 3:(a)], where $Y$-axis shows the suboptimality $P(\tilde{x}_t) - P(x_\star)$ and $X$-axis shows data passes or running time. For rcv1, we set $\lambda_2$ and $\lambda_1$ to be different values, which are noted along with the figures. Figure 3 shows that OPDA-FM stably outperforms Proximal-SVRG, especially when the strong convexity coefficient $\mu$ is smaller; Figure 4 shows that the advantage of OPDA-FM over Proximal-SVRG is more prominent when the $L_1$ regularization is stronger.

For OPDA-QN, we set $M = 5$ as the memory size of curvature triples/pairs. The updating frequency of $H_k$ also is also set to be $M$. We plot the suboptimality against the datapasses and the running time in Figure 5. The sampling strategies for $\Xi_k$ are noted in the figure, like Guassian sampling as **gauss**, previous directions as **prev**, and

$\Xi_k$ being an identity matrix as **BFGS**. All OPDA-QN algorithms run considerably faster than Proximal-SVRG, by the advantages of second-order information. For low-dimensional data like covtype, OPDA-gauss, OPDA-prev strategies perform slightly different, but both outperform OPDA-BFGS, showing the effectiveness of the sketching technique. For high-dimensional data like rcv1, OPDA-BFGS perform faster than OPDA-prev, both in measure of data passes or running time, and they all outperform OPDA-gauss, since they consume less computations for calculating $d_k = H_k v_k$; and the running time per iteration of OPDA-BFGS is less than that of OPDA-prev. The acceleration of OPDA-QN in terms of data passes is very critical for modern machine learning applications. In the past decades, the computation power grew faster than the memory bandwidth, and this trend benefits OPDA-QN more, since QN type algorithms generally visit data less frequently by consuming more computations per iteration.

### Deep Learning

We also conducted experiments with sparse convolutional neural networks for demonstrating nonconvex optimization efficiency and potential application in memory limited deep learning. For the consistency with other algorithms in the area, we use the stochastic gradient without variance reduction or quasi-Newton methods, so that $d_k = v_k = \diamondsuit f_k(x_{k-1})$. The algorithm is noted as OPDA-SGD. The network has three convolutional layers of $5 \times 5 \times 16$, $3 \times 3 \times 32$ and $3 \times 3 \times 48$ (kernel height, kernel width, output channels), three pooling layers of $(2, 2)$ (size, stride), a fully connected layer and a softmax loss layer. We use the **CIFAR10** dataset. The parameter of the network has $L_1$ regularization, whose coefficient $\lambda$ is set to be $[10, \cdots, 10/2^7]$, and noted in the figure. We see that OPDA-SGD converges considerably faster than Proximal-SGD, as shown in Figure 6, especially when the $L_1$ regularization is stronger. This agrees with our intuition, since OPDA is specifically designed to tackle the strong non-differentiability.

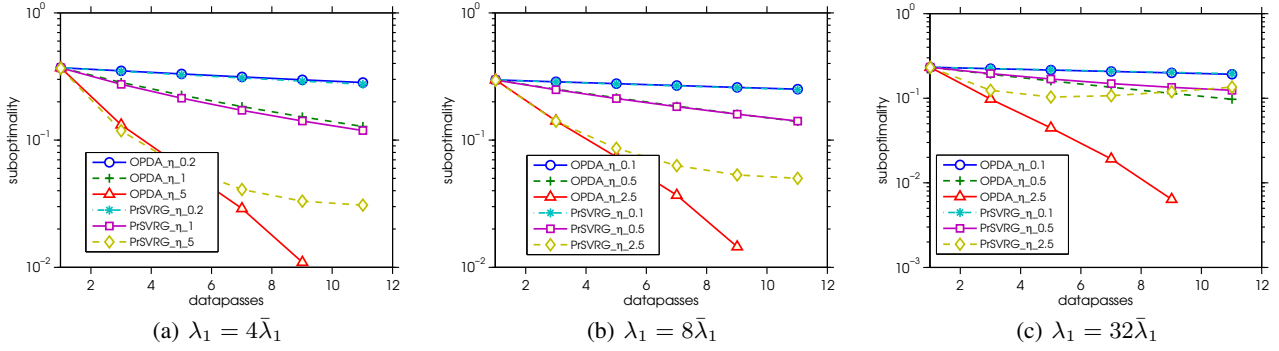

Figure 4: Performance of OPDA-FM for different $L_1$ regularization on rcv1 dataset, $\bar{\lambda}_1 = 10^{-5} \sqrt[4]{D}/\sqrt[4]{N}$.

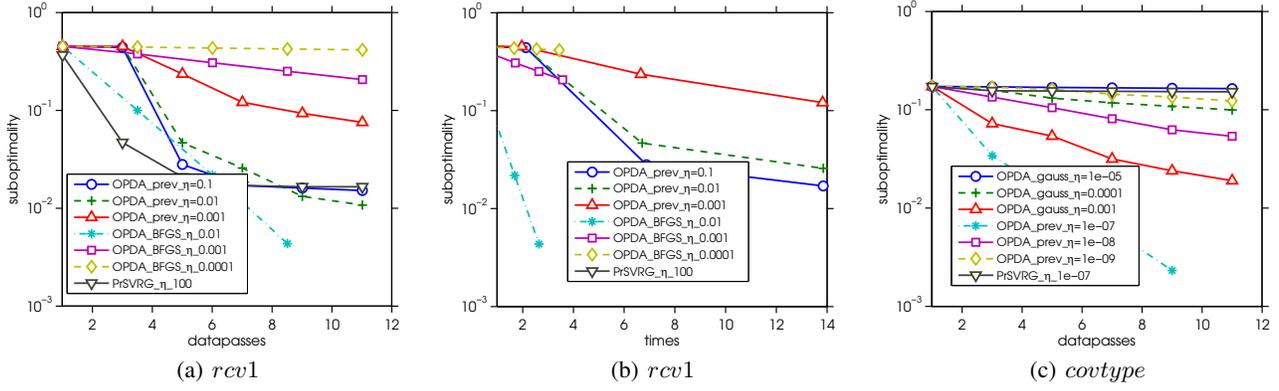

Figure 5: Performance of OPDA-QN with different strategies and different datasets.

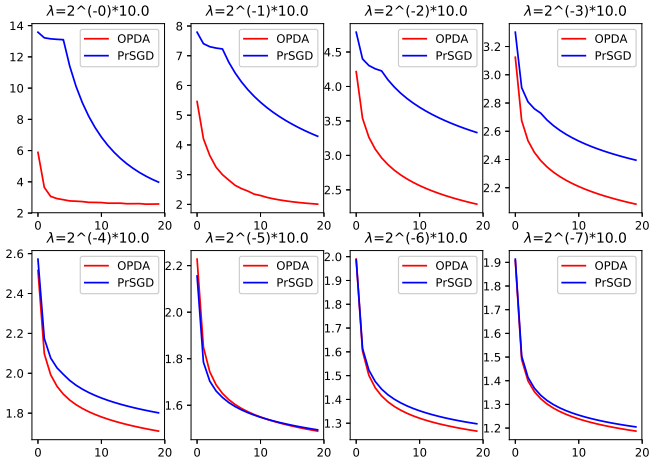

Figure 6: The performance of OPDA-SGD on CIFAR10. X-axis: datapasses; Y:axis: objective function.

In conclusion, as we see, although by the orthant-wise nature of OPDA, a large proportion of elements of the descent direction $p$ and the updated parameter $x - \eta p$ are forced to zero during alignment, making the actual descent speed slower, OPDA-FM and OPDA-QN still converge much faster than state-of-the-art proximal algorithms, with the same stepsize. This proves that our proposed alignment operator does calibrate the direction to be a better one toward the optimum, making the algorithm considerably more efficient, with only negligible extra arithmetic operations consumed. In addition, OPDA-QN, with or without sketching, outperforms OPDA-FM under proper settings.

## Conclusion

We propose OPDA as a good substitute of proximal algorithms to efficiently train $L_1$ regularized sparse models. We apply SVRG to initialize the descent direction, which can be calibrated by quasi-Newton methods (OPDA-QN). We propose a novel alignment operator to encourage the differentiability of the $L_1$ regularization. The parameter therefore cross each orthant passively during optimization. We prove a linear convergence rate of the algorithm with strong convexity and smoothness, and the experiments show that OPDA stably and significantly outperforms proximal algorithms on both convex and nonconvex problems.

**Proof of Lemma 8.**
Consider the function
$$\phi_k(x) = f_k(x) - f_k(x_\star) - \nabla f_k(x_\star)^\top (x - x_\star)$$
It is straightforward to check that $\nabla \phi_k(x_\star) = 0$, hence
$$\min_x \phi_k(x) = \phi_k(x_\star) = 0.$$
Since $\phi_k(x)$ is Lipschitz continuous with constant $L_i$, we have
$$\frac{1}{2L}||\nabla \phi_k(x)||^2 \leq \phi_k(x) - \min_y \phi_k(y)$$
$$\leq \phi_k(x) - \phi_k(x_\star) = \phi_k(x)$$
This implies
$$||\nabla f_k(x) - \nabla f_k(x_\star)||^2$$
$$\leq 2L[f_k(x) - f_k(x_\star) - \nabla f_k(x_\star)^\top (x - x_\star)],$$
taking expectation with respect to data set $S_k$, we get
$$\mathbb{E}||\nabla f_k(x) - \nabla f_k(x_\star)||^2$$
$$\leq F(x) - F(x_\star) - \nabla F(x_\star)^\top (x - x_\star).$$
By the optimality of $x_\star$, i.e.,
$$x_\star = \operatorname*{argmin}_x P(x) = \operatorname*{argmin}_x \{F(x) + R(x)\},$$
there exist $\xi \in \partial R(x_\star)$ such that $\nabla F(x_\star) + \xi = 0$. Therefore
$$F(x) - F(x_\star) - \nabla F(x_\star)(x - x_\star)$$
$$= F(x) - F(x_\star) + \xi^\top (x - x_\star)$$
$$\leq F(x) - F(x_\star) + R(x) - R(x_\star) = P(x) - P(x_\star),$$
where the inequality is due to convexity of $R(x)$. This proves the desired result.

**Proof of Lemma 9.**
Conditioned on $x_{k-1}$, we take expectation with respect to $S_k$ to obtain
$$\mathbb{E}[\nabla f(x_{k-1})] = \nabla F(x_{k-1})$$
Similarly we have $\mathbb{E}[\nabla f_k(\tilde{x})] = \nabla F(\tilde{x})$, and
$$\mathbb{E}[v_k] = \mathbb{E}[\nabla f_k(x_{k-1})] - \mathbb{E}[\nabla f_k(\tilde{x})] + \nabla F(x_{k-1}),$$
therefore, $\mathbb{E}[v_k] = \nabla F(x_{k-1})$. Here we introduce two inequality as
$$\mathbb{E}||\xi - \mathbb{E}\xi||^2 \leq \mathbb{E}||\xi||^2, \quad \forall \xi \in \mathbb{R}^D$$
and $||a + b||^2 \leq 2||a||^2 + 2||b||^2$. Then we can bound the variance, we have
$$\mathbb{E}[v_k^2] = 2\mathbb{E}||\nabla f_k(x_{k-1})||^2 + 2\mathbb{E}||\nabla f_k(\tilde{x}) - \nabla F(\tilde{x})||^2$$
$$\leq \mathbb{E}||\nabla f_k(x_{k-1})||^2 + \mathbb{E}||\nabla f_k(\tilde{x})||^2$$
$$\leq 4L[P(x_{k-1}) - P(x_\star) + P(\tilde{x}) - P(x_\star)]$$

**Proof of Lemma 12.**
We take each elements of variables into discussion. (1) if $z_i = 0$ or $\sigma(x_i) = \sigma(y_i) = \sigma(z_i)$, there is
$$\phi_i(x; z; \eta\lambda) = \sigma(x_i) \max(|x_i| - \eta\lambda, 0) = (prox_{\eta R}(x))_i, \quad (35)$$
reducing to a case included in Lemma 10; (2) else, there is $z_i \neq 0$, without loss of generality, we assume $\sigma(x_i) = \sigma(z_i)$ and $\sigma(y_i) \neq \sigma(z_i)$, so
$$\phi_i(y; z; \eta\lambda) = 0 = \phi_i(0; z; \eta\lambda),$$
then
$$||\phi_i(x; z; \eta\lambda) - \phi_i(0; z; \eta\lambda)|| \leq ||x_i - 0|| \leq ||x_i - y_i||; \quad (36)$$
(3) else, there is $z_i \neq 0$, $\sigma(x_i) \neq \sigma(z_i)$ and $\sigma(y_i) \neq \sigma(z_i)$, so
$$\phi_i(x; z; \eta\lambda) - \phi_i(y; z; \eta\lambda) = 0.$$
Combining the inequalities for all elements, we get the proof.

**Proof of Lemma 13.**
We define a subgradient of $R$ as $\xi \in \partial R(x^+)$, for each element $\xi_i$, $1 \leq i \leq D$, (1) if $\sigma(x_i^-)\sigma(x_i) \neq -1$, we can obtain $x_i^+ = \sigma(x_i^-) \max(|x_i^-| - \eta\lambda, 0)$, which is exactly the optima of
$$x_i^+ = \operatorname*{argmin}_{y_i} \frac{1}{2}||y_i - (x_i - \eta p_i)||^2 + \eta R(y_i), \quad (37)$$
then by the optimality property of $x^+$ and the definition $q_i = (x_i - x_i^+)/\eta$, it holds
$$x_i^+ - (x_i - \eta p_i) + \eta\xi_i = 0 \Rightarrow \xi_i = q_i - p_i, \quad (38)$$
(2) else, there is $\sigma(x_i^-)\sigma(x_i) = -1$, then $x_i^+ = 0$, leading to $\xi_i = 0$, which is rewritten as $\xi_i = q_i - q_i$. Combining condition (1) and (2) for each elements, we can conclude that there exist a sub-gradient $\xi = q - g \in \partial R(x^+)$.

The function $P(x)$ is lower bounded by the strong convexity of $F$ and $R$, for any $x, y \in \mathbb{R}^D$,
$$P(y) = F(y) + R(y) \geq F(x) + \nabla F(x)^\top (y - x) \quad (39)$$
$$+ \mu||y - x||^2 + R(x^+) + \xi^\top (y - x^+) \quad (40)$$
The term $F(x)$ is further bounded by its smoothness,
$$F(x) \geq F(x^+) - \nabla F(x)^\top (x^+ - x) - \frac{L}{2}||x^+ - x||^2. \quad (41)$$
Therefore there is
$$P(y) \geq F(x^+) - \nabla F(x)^\top (x^+ - x) - \frac{L}{2}||x^+ - x||^2$$
$$+ \nabla F(x)^\top (y - x) + \frac{\mu}{2}||y - x||^2 + R(x^+) + \xi^\top (y - x^+)$$
$$\geq P(x^+) - \nabla F(x)^\top (x^+ - x) - \frac{L}{2}\eta^2||q||^2$$
$$+ \nabla F(x)^\top (y - x) + \frac{\mu}{2}||y - x||^2 + \xi^\top (y - x^+) \quad (42)$$
where in the last inequality we used and $x^+ - x = -\eta q$. For the production terms on the right-hand side, we have
$$-\nabla F(x)^\top (x^+ - x) + \nabla F(x)^\top (y - x) + \xi^\top (y - x^+)$$
$$= \nabla F(x)^\top (y - x^+) + (q - g)^\top (y - x^+)$$
$$= q^\top (y - x^+) + (g - \nabla F(x))^\top (x^+ - y)$$
$$= q^\top (y - x + x - x^+) + \Delta^\top (x^+ - y)$$
$$= q^\top (y - x) + \eta||q||^2 + \Delta^\top (x^+ - y) \quad (43)$$

where we used $\xi = q - g$, $\Delta = g - \nabla F(x)$ and $x - x^+ = \eta q$ in the first, third, last equalities respectively. Substituting Eq.(42) to Eq.(43), we obtain Eq.(31).

**Proof of Lemma 14.**

We apply the Cauchy-Schwarz inequality for each element $\Delta_i(x_i^+ - y_i), \forall 1 \leq i \leq D$,

$$-\Delta_i(x_i^+ - y_i) = -\Delta_i(x_i^+ - \bar{x}_i) - \Delta_i(\bar{x}_i - y_i)$$
$$\leq ||\Delta_i||||(x_i^+ - \bar{x}_i)|| - \Delta_i(\bar{x}_i - y_i) \quad (44)$$

Then we apply Lemma 10 if $\sigma(x_i^-)\sigma(x_i) \neq -1$, and apply Lemma 12 if $\sigma(x_i^-)\sigma(x_i) = -1$, we have $||x_i^+ - \bar{x}_i|| \leq \eta||\Delta_i||$ by the definition of $g$ and $\Delta$. Combining them, we get the first inequality.

**Proof of Lemma 12.**

We prove the theorem by analyse how the Euclidean distance between $x_k$ and $x_\star$ changes over iterations. We define an auxiliary function

$$\mathcal{L}_k = ||x_k - x_\star||^2 - ||x_{k-1} - x_\star||^2, \quad (45)$$

since $x_k = x_{k-1} - \eta q_k$, then it holds

$$\mathcal{L}_k = -2\eta q_k^\top(x_{k-1} - x_\star) + \eta^2||q_k||^2. \quad (46)$$

Then we apply Lemma 13 with substitution that $x = x_{k-1}$, $v = v_k$, $x^+ = x_k$, $q = q_k$, and $y = x_\star$,

$$-q_k^\top(x_{k-1} - x_\star) + \frac{\eta}{2}||q_k||^2 \leq P(x_\star) - P(x_k)$$
$$-\frac{\mu}{2}||x_{k-1} - x_\star||^2 - \Delta_k^\top(x_k - x_\star), \quad (47)$$

Denoting $Q_k = Q(x_k)$ and combining Eqs.(46;47) together,

$$\mathcal{L}_k \leq -\eta\mu||x_{k-1} - x_\star||^2 - 2\eta Q_k - 2\eta \Delta_k^\top(x_k - x_\star)$$
$$\leq -2\eta Q_k - 2\eta \Delta_k^\top(x_k - x_\star). \quad (48)$$

Next we bound the term $-2\eta \Delta_k^\top(x_k - x_\star)$. We define an auxiliary variable $\bar{x}_k \in \mathbb{R}^D$ as

$$\bar{x}_k = prox_{\eta R}(x_{k-1} - \eta \nabla F(x_{k-1})), \quad (49)$$

which is dependent of $x_{k-1}$ but independent of random set $\mathcal{S}_k$, then

$$-2\eta \Delta_k^\top(x_k - x_\star) \leq 2\eta^2||\Delta_k^\top||^2 - 2\eta \Delta_k^\top(\bar{x}_k - x_\star). \quad (50)$$

Combining with Eq.(48), we get

$$\mathcal{L}_k \leq -2\eta Q_k + 2\eta^2||\Delta_k||^2 - 2\eta \Delta_k^\top(\bar{x}_k - x_\star) \quad (51)$$

Now we take expectation on both sides of the above inequality with respect to $\mathcal{S}_k$ to obtain

$$\mathbb{E}[\mathcal{L}_k] \leq -2\eta\mathbb{E}[Q_k] + 2\eta^2\mathbb{E}||\Delta_k||^2 - 2\eta\mathbb{E}[\Delta_k^\top(\bar{x}_k - x_\star)] \quad (52)$$

Recalling that $g_k$ and $x_{k-1}$ are both symmetrically distributed across zero point, they are with zero means, then

$$\mathbb{E}[\Delta_k] = \mathbb{E}[g_k] - \mathbb{E}[\nabla F(x_{k-1})] = 0. \quad (53)$$

We note that both $\bar{x}_k$ and $x_\star$ are independent of the random set $\mathcal{S}_k$, therefore they are not correlated with the term $\Delta_k$.

$$\mathbb{E}[\Delta_k^\top(\bar{x}_k - x_\star)] = \mathbb{E}[\Delta_k]^\top(\bar{x}_k - x_\star) = 0. \quad (54)$$

Recalling the definition of $p_k$ and Lemma 14, there is

$$\mathbb{E}||p_k||^2 \leq \mathbb{E}||H_k v_k||^2 \leq \Gamma^2 \mathbb{E}||v_k||^2. \quad (55)$$

by the definition of spectral norm. We write the term

$$\mathbb{E}||\Delta_k||^2 = \mathbb{E}||g_k - \nabla F(x_{k-1})||^2, \quad (56)$$

and apply Lemma 14 that $||g_k||^2 \leq ||p_k||^2$, then

$$\mathbb{E}||\Delta_k||^2 = \mathbb{E}[||g_k||^2 - 2g_k^\top \nabla F(x_{k-1}) + ||\nabla F(x_{k-1})||^2]$$
$$\leq \mathbb{E}[||p_k||^2 - 2(p_k + \xi_k)^\top \nabla F(x_{k-1}) + ||\nabla F(x_{k-1})||^2]. \quad (57)$$

where $\xi_k = g_k - p_k$ is dependent of data subset $\mathcal{S}_k$, therefore is independent of $\nabla F(x_{k-1})$. Recalling that $\mathbb{E}[\xi_k] = 0$, so we have $\mathbb{E}[\xi_k^\top \nabla F(x_{k-1})] = 0$,

$$\mathbb{E}||\Delta_k||^2 \leq \mathbb{E}[||p_k||^2 - 2p_k^\top \nabla F(x_{k-1}) + ||\nabla F(x_{k-1})||^2]. \quad (58)$$

Since $p_k$ and $\nabla F(x_{k-1})$ are independent with each other and have zero means, which leads to $\mathbb{E}[p_k^\top \nabla F(x_{k-1})] = 0$, substitute this and Eq.(55) into term $||\Delta_k||^2$, then there is

$$\mathbb{E}||\Delta_k||^2 \leq \Gamma^2 \mathbb{E}||v_k||^2 + \mathbb{E}||\nabla F(x_{k-1})||^2$$
$$\leq (4L\Gamma^2 + 2L)Q_{k-1} + 4L\Gamma^2 Q(\tilde{x}), \quad (59)$$

where we apply Lemma 1 to bound the term $||v_k||$ and $||\nabla F(x_{k-1})||$. Substituting Eqs.(54;59) into Eq.(52),

$$\mathcal{L}_k \leq -2\eta Q_k + (8\Gamma^2 + 4)L\eta^2 Q_{k-1} + 8L\Gamma^2\eta^2 Q(\tilde{x}). \quad (60)$$

We take the optimization trajectory from with $(t-1)m$-th step for analysis, where the reference point $\tilde{x}_{t-1}$ were used, then

$$x_0 = \tilde{x} = \tilde{x}_{t-1}, \quad \tilde{x}_t = \frac{1}{m} \cdot \sum_{k=1}^{m} x_k \quad (61)$$

By combining the inequalities in Eq.(60), at iterations $k = 1, \cdots, m$, we obtain

$$\mathbb{E}||x_m - x_\star||^2 + 2\eta Q_m + \alpha \sum_{k=1}^{m-1} Q_k$$
$$= \mathbb{E}||x_m - x_\star||^2 + \alpha \sum_{k=1}^{m} Q_k + (8L\Gamma^2 + 4L)\eta^2$$
$$\leq ||x_0 - x_\star||^2 + \beta[Q(x_0) + mQ(\tilde{x})] \quad (62)$$

where $\alpha = 2\eta(1 - 4L\Gamma^2\eta - 2L\eta)$ and $\beta = 8L\Gamma^2\eta^2$. Notice that $2\eta(1 - 4L\Gamma^2\eta) < 2\eta$ and $x_0 = \tilde{x}_{t-1}$, so we have

$$\alpha \sum_{k=1}^{m} Q_k \leq ||\tilde{x}_{t-1} - x_\star||^2 + \beta(m+1)Q(x_0). \quad (63)$$

By the strong convexity of $P$ and definition of $\tilde{x}_t$, we have

$$P(\tilde{x}_t) \leq \frac{1}{m} \sum_{k=1}^{m} P(x_k), ||\tilde{x}_{t-1} - x_\star||^2 \leq \frac{2}{\mu} \cdot Q(\tilde{x}_{t-1}), \quad (64)$$

Substituting it back, we have

$$\alpha m Q(\tilde{x}_t) \leq (\frac{2}{\mu} + \beta(m+1))Q(\tilde{x}_{t-1}) \quad (65)$$

Recall that $\eta < 1/(4L\Gamma^2 + 2L)$, there is $\alpha > 0$, dividing both sides of the above inequality by $\alpha m$, we arrive at

$$Q(\tilde{x}_t) \leq (\frac{2}{\mu\alpha m} + \frac{\beta(m+1)}{\alpha m})Q(\tilde{x}_{t-1}) \quad (66)$$

Define $\rho = \frac{2}{\mu\alpha m} + \frac{\beta(m+1)}{\alpha m}$, and apply the above inequality recursively, we get

$$Q(\tilde{x}_t) \leq \rho^t Q(\tilde{x}_0), \quad (67)$$

which is our main theorem.

Now, we proceed to the analysis on nonconvex problems. Comparatively, the criterion of the analysis for nonconvex but smooth problems is clear, since the norm of gradients $||\nabla F(x)||_2$ provides a good measure of stationarity. But for OPDA on nonsmooth problems, we have to define a proximal mapping as (Nesterov 2013). First, we define

$$\bar{p}_k = \pi(H_k \nabla F(x_{k-1}); \diamondsuit f(x_{k-1})).$$

Then the gradient mappings of OPDA is defined as

$$\mathcal{G}(x_k) = \frac{1}{\eta} \left[ \phi(x_{k-1} - \eta \bar{p}_k, x_{k-1}, \eta\lambda) - x_{k-1} \right].$$

**Lemma 18.** *Under the condition of Assumption 1 but without the convexity assumption, there is*

$$\mathbb{E}[||v_k - \nabla F(\tilde{x})||^2] \leq \frac{L^2}{M}||x_k - \tilde{x}||^2.$$

**Lemma 19.** *For the conditions and notations in Lemma 13, but without the convexity assumption, if we define $y = \phi(x - \eta g; x; \eta\lambda)$, then for all $z \in \mathbb{R}^D$, there is*

$$P(y) \leq P(z) + (y-z)^\top (\nabla F(x) - g) + (\frac{L}{2} - \frac{1}{2\eta})||y - x||^2$$
$$+ (\frac{L}{2} + \frac{1}{2\eta})||z - x||^2 - \frac{1}{2\eta}||y - z||^2.$$

**Remark 9.** The lemma is essentially a deformation of the smoothness inequality that

$$F(y) \leq F(x) + \nabla F(x)^\top (y - x) + \frac{L}{2}||y - x||^2,$$
$$F(x) \leq F(z) + \nabla F(x)^\top (x - z) + \frac{L}{2}||x - z||^2,$$

and the non-expansive property of the alignment operator.

**Theorem 20.** *For loss functions of $L$-smoothness, which may be nonconvex, we suppose that Lemma 11 holds for the quasi-Newton update matrix, then by utilizing Algorithm 1, if we set the subsampling size as $M = N^{2/3}$, the inner iterations as $m = N/M$ and the stepsize as $\eta = M/(4L\Gamma N)$, and the maximum outer iterations $T$ to be a multiple of $m$, then there is*

$$\mathbb{E}[||\mathcal{G}(\tilde{x}_T)||^2] \leq 16\frac{L^3\Gamma^3 N}{T}\left(P(x_0) - P(x_\star)\right). \quad (68)$$

**Remark 10.** The convergence rate of OPDA for nonconvex function has a similar form with proximal-SVRG (Reddi et al. 2016). The difference lies in the variance of quasi-Newton directions.

**Proof of Theorem 20.**

*Proof.* We define several auxiliary variables for analysis,

$$\nu_k = x_k - x_{k-1}, \quad \bar{\nu}_k = \bar{x}_k - x_{k-1}, \quad \underline{\nu}_k = x_k - \bar{x}_k$$

Then we apply Lemma 19, and substitute $y \leftarrow \bar{x}_k$, $z \leftarrow x_{k-1}$ and $g \leftarrow \bar{g}_k$, to get

$$\mathbb{E}[P(\bar{x}_k)] \leq \mathbb{E}[P(x_{k-1}) + \bar{\nu}_k^\top (\nabla F(x_{k-1}) - \bar{g}_k) + (\frac{L}{2} - \frac{1}{2\eta})||\bar{\nu}_k||^2 - \frac{1}{2\eta}||\bar{\nu}_k||^2].$$

By applying Lemma 19, and substituting $y = x_k, z = \bar{x}_k$, $g = g_k$, we also have a similar result for $x_k$ as

$$\mathbb{E}[P(x_k)] \leq \mathbb{E}[P(\bar{x}_k) + \underline{\nu}_k^\top (\nabla F(x_{k-1}) - g_k) + (\frac{L}{2} - \frac{1}{2\eta})||\nu_k||^2 + (\frac{L}{2} + \frac{1}{2\eta})||\bar{\nu}_k||^2 - \frac{1}{2\eta}||\underline{\nu}_k||^2].$$

Combining the two inequalities above, we get

$$\mathbb{E}[P(x_k)] \leq \mathbb{E}[P(x_{k-1}) + \bar{\nu}_k^\top (\nabla F(x_{k-1}) - \bar{g}_k) + \underline{\nu}_k^\top (\nabla F(x_{k-1}) - g_k) + (L - \frac{1}{2\eta})||\bar{\nu}_k||^2 + (\frac{L}{2} - \frac{1}{2\eta})||\nu_k||^2 - \frac{1}{2\eta}||\underline{\nu}_k||^2],$$

we can bound the inner product terms by

$$\underline{\nu}_k^\top (\nabla F(x_{k-1}) - g_k) \leq \frac{1}{2\eta}\mathbb{E}[||\underline{\nu}_k||^2] + \frac{\eta}{2}||\nabla F(x_{k-1}) - g_k||^2$$

$$\bar{\nu}_k^\top (\nabla F(x_{k-1}) - \bar{g}_k) \leq \frac{1}{4\eta}\mathbb{E}[||\bar{\nu}_k||^2] + \eta||\nabla F(x_{k-1}) - \bar{g}_k||^2$$

by the Cauchy-Schwarz inequality, and the smoothness of $F(x)$. By the non-expansive property in Lemma 12,

$$\mathbb{E}[||\underline{\nu}_k||^2] \leq \mathbb{E}[||\bar{p}_k - p_k||^2] \leq \mathbb{E}[||H_k \nabla F(x_{k-1}) - H_k v_k||^2]$$
$$\leq \Gamma^2 \mathbb{E}[||\nabla F(x_{k-1}) - v_k||^2] \leq \frac{\Gamma^2 L^2}{M}||x_{k-1} - \tilde{x}||^2].$$

We define $\Delta_k = g_k - \nabla F(x_{k-1})$. First, by the definition of $p_k$ and Lemma 11, there is

$$\mathbb{E}||g_k||^2 \leq \mathbb{E}||p_k||^2 \leq \mathbb{E}||H_k v_k||^2, \quad (69)$$

by the definition of spectral norm. We write the term $\mathbb{E}||\Delta_k||^2$ with Lemma 14 that $||g_k||^2 \leq ||p_k||^2$, then

$$\mathbb{E}||\Delta_k||^2 = \mathbb{E}[||g_k||^2 - 2g_k^\top \nabla F(x_{k-1}) + ||\nabla F(x_{k-1})||^2]$$
$$\leq \mathbb{E}[||p_k||^2 - 2(p_k + \xi_k)^\top \nabla F(x_{k-1}) + ||\nabla F(x_{k-1})||^2]$$
$$\leq \mathbb{E}[||H_k v_k||^2 - 2(H_k v_k)^\top \nabla F(x_{k-1}) + ||\nabla F(x_{k-1})||^2]$$

where the second inequality is by that $\xi_k = g_k - p_k$ is dependent of data subset $S_k$, therefore is independent of $\nabla F(x_{k-1})$, so $\mathbb{E}[\xi_k^\top \nabla F(x_{k-1})] = 0$.

$$\mathbb{E}||\Delta_k||^2 \leq \mathbb{E}[||H_k v_k - \nabla F(x_{k-1})||^2]$$
$$\leq \mathbb{E}[||H_k (v_k - \nabla F(x_{k-1})) + (H_k - I)\nabla F(x_{k-1})||^2)]$$
$$\leq \mathbb{E}[2\Gamma^2 ||v_k - \nabla F(x_{k-1})||^2 + 2(\Gamma + 1)^2 ||\nabla F(x_{k-1})||^2]$$

And the inequality is by Lemma 11 to bound the term $||v_k||$ and $||\nabla F(x_{k-1})||$. By setting the upper bound parameter $\Gamma \leftarrow \Gamma + 1$ we can get

$$\mathbb{E}||\Delta_k||^2 \leq 2\Gamma^2 \mathbb{E}[||v_k - \nabla F(x_{k-1})||^2 + ||\nabla F(x_{k-1})||^2]$$
$$\leq 2\Gamma^2(\frac{L^2}{M}||x_{k-1} - \tilde{x}||^2] + L^2||x_{k-1} - x_\star||^2)$$

Substituting this back, we have

$$\underline{\nu}_k^\top(\nabla F(x_{k-1}) - g_k) \leq \eta\Gamma^2(\frac{L^2}{M}||x_{k-1} - \tilde{x}||^2]$$
$$+ L^2||x_{k-1} - x_\star||^2),$$

and we have similar results as

$$\bar{\nu}_k^\top(\nabla F(x_{k-1}) - \bar{g}_k) \leq \frac{1}{4\eta}\mathbb{E}[||\bar{\nu}_k||^2] + \eta\Gamma^2 L^2||x_{k-1} - x_\star||^2.$$

Substituting these two inequalities back,

$$\mathbb{E}[P(x_k)] \leq \mathbb{E}[P(x_{k-1}) + (L - \frac{1}{4\eta})||\bar{\nu}_k||^2 + (\frac{L}{2} - \frac{1}{2\eta})||\nu_k||^2$$
$$+ \frac{\eta\Gamma^2 L^2}{M}||x_{k-1} - \tilde{x}||^2 + 2\eta\Gamma^2 L^2 ||x_{k-1} - x_\star||^2,$$

We analyse the distance between the current estimation with the reference point $\tilde{x}$, to be specific,

$$\mathbb{E}[||x_k - \tilde{x}||^2] = \mathbb{E}[||\nu_k + x_{k-1} - \tilde{x}||^2]$$
$$= \mathbb{E}[||\nu_k||^2 + ||x_{k-1} - \tilde{x}||^2 + 2\nu_k^\top(x_{k-1} - \tilde{x})],$$

we also have

$$\mathbb{E}[||x_k - x_\star||^2] = \mathbb{E}[||\nu_k||^2 + ||x_{k-1} - x_\star||^2 + 2\nu_k^\top(x_{k-1} - x_\star)]$$

By constructing a variable $\mathcal{L}_k$ as

$$\mathcal{L}_k = \mathbb{E}[P(x_k) + c_k||x_k - \tilde{x}||^2 + b_k||x_k - x_\star||^2,$$

where $c_k$ and $b_k$ are constants that

$$c_{k-1} = c_k(1+\beta) + \eta\Gamma^2 L^2/M,$$
$$b_{k-1} = b_k(1+\beta) + 2\eta\Gamma^2 L^2$$

Combining these inequalities together, we have

$$\mathcal{L}_k = \mathbb{E}[P(x_k) + c_k||x_k - \tilde{x}||^2] + b_k||x_k - x_\star||^2$$
$$= \mathbb{E}[P(x_k)] + c_k\mathbb{E}[||\nu_k||^2 + ||x_{k-1} - \tilde{x}||^2 + 2\nu_k^\top(x_{k-1} - \tilde{x})]$$
$$+ b_k\mathbb{E}[||\nu_k||^2 + ||x_{k-1} - x_\star||^2 + 2\nu_k^\top(x_{k-1} - x_\star)]$$
$$\leq \mathbb{E}[P(x_k)] + \mathbb{E}[(b_k + c_k)(1+\frac{1}{\beta})||\nu_k||^2 + c_k(1+\beta)||x_{k-1} - \tilde{x}||^2$$
$$\leq \mathbb{E}[P(x_{k-1}) + (L - \frac{1}{4\eta})||\bar{\nu}_k||^2 + [c_k(1+\frac{1}{\beta})$$
$$+ (\frac{L}{2} - \frac{1}{2\eta})]||\nu_k||^2 + [c_k(1+\beta) + \frac{\eta L^2\Gamma^2}{2M}]||x_{k-1} - \tilde{x}||^2]$$
$$+ [b_k(1+\beta) + 2\eta\Gamma^2 L^2 ||x_{k-1} - x_\star||^2]$$

By apply the definition of $b_k$ and $c_k$ we have

$$\mathcal{L}_k \leq \mathbb{E}[P(x_{k-1}) + (L - \frac{1}{4\eta})||\bar{\nu}_k||^2 + [c_k(1+\beta)$$
$$+ \frac{\eta L^2\Gamma^2}{2M}]||x_{k-1} - \tilde{x}||^2 + [b_k(1+\beta) + 2\eta\Gamma^2 L^2]||x_{k-1} - x_\star||^2],$$

where the third inequality is due to the following inequality

$$(b_k + c_k)(1 + \frac{1}{\beta}) + \frac{L}{2} \leq \frac{1}{2\eta}, \quad (70)$$

which can be recursively proved. First set $\beta = M/N$, $m = int(N/M)$ and $c_m = 0$, then since $c_{k-1} = c_k(1+\beta) + \eta L^2\Gamma^2/M$ then

$$c_k = \frac{\eta L^2\Gamma^2}{M}\frac{(1+\beta)^{m-k-1} - 1}{\beta}$$
$$= \frac{\eta L^2\Gamma^2 N}{M^2}[(1 + \frac{M}{N})^{m-k-1} - 1]$$
$$\leq \eta L^2\Gamma^2 N[(1 + \frac{M}{N})^{N/M} - 1]/M^2$$
$$\leq \eta L^2\Gamma^2 N(e-1)/M^2$$

where the first inequality is due to $m = N/M$, and the second inequality is by the definition of $e$, the Euler's number. Similarly for $b_k$, we have

$$b_k \leq \eta L^2\Gamma^2 N(e-1)/M.$$

Combining two inequalities above, and substituting $\eta < M/(4L\Gamma N)$ and $\Gamma \geq 1$, there is

$$(b_k + c_k)(1 + \frac{1}{\beta}) + \frac{L}{2}$$
$$\leq L\Gamma(e-1)N\frac{M}{4N}\frac{1+M}{M^2}(1 + \frac{N}{M}) + \frac{L}{2}$$
$$\leq L\Gamma(e-1)\frac{N}{2M} + \frac{L\Gamma}{2}$$
$$\leq \frac{3L\Gamma}{2}\frac{N}{M} \leq 2L\Gamma\frac{N}{M} \leq \frac{1}{2\eta}$$

Substituting this back, we have

$$\mathcal{L}_k \leq \mathcal{L}_{k-1} + (L - \frac{1}{4\eta})||\bar{\nu}_k||^2.$$

By summing the inequalities from $k = mt$ to $k = m(t+1)$, $b_m = c_m = 0$, we have

$$\mathcal{L}_{m(t+1)} \leq \mathcal{L}_{mt} + \sum_{k=mt}^{m(t+1)}(L - \frac{1}{4\eta})\mathbb{E}[||\bar{x}_k - x_{k-1}||^2].$$

And by the definition that $x_{mt} = \tilde{x}$ in this outer iteration, we have

$$\mathbb{E}[P(\tilde{x}_{t+1}) + b_0||\tilde{x}_{t+1} - x_\star||^2]$$
$$\leq \mathbb{E}[P(\tilde{x}_t)] + \sum_{k=mt}^{m(t+1)}(L - \frac{1}{4\eta})\mathbb{E}[||\bar{x}_k - x_{k-1}||^2].$$

And by summing up all the outer iterations,

$$\sum_{k=0}^{m(t+1)}(\frac{1}{4\eta} - L)\mathbb{E}[||\bar{\nu}_k||^2] \leq P(x_0) - \mathbb{E}[P(x_{m(t+1)})]$$
$$\leq P(x_0) - P(x_\star).$$

By recalling that $\mathcal{G}(x_k) = \bar{\nu}_k/\eta$, we have

$$\mathbb{E}[||\mathcal{G}(\tilde{x}_T)||^2] \leq 32L^3\Gamma^3 N^3 \left(P(x_0) - P(x_\star)\right)/(M^3 T),$$

By substituting $M = N^{2/3}$, we conclude the main theorem. $\square$